\documentclass[11pt]{article}

\usepackage[utf8]{inputenc}
\usepackage[T1]{fontenc}
\usepackage{lmodern}
\usepackage[margin=1in]{geometry}
\usepackage{amsmath,amssymb}
\usepackage{booktabs}
\usepackage{array}
\usepackage{siunitx}
\usepackage{xcolor}
\usepackage{microtype}
\usepackage[numbers,sort&compress]{natbib}
\usepackage[hidelinks]{hyperref}
\usepackage{enumitem}
\usepackage{graphicx}
\usepackage{chessboard}
\usepackage{xskak}
\usepackage{pgfplots}
\pgfplotsset{compat=1.18}

\newcommand{\sfzero}{$\text{SF}\!\approx\!0$}
\newcommand{\eval}{\textit{eval}}

\title{\textbf{Engine-Equal, Human-Unequal:\\
A Reproducible Outcome Skew in\\
Engine-Assessed Equal Chess Positions}}

\author{Jesung Park\\
  \small Gamakon\\
  \small \texttt{jesung@gamakon.ai}}

\date{\today\\[4pt]
  \small Preprint --- standalone observational result. Not peer reviewed.}

\hypersetup{
  pdftitle={Engine-Equal, Human-Unequal: A Reproducible Outcome Skew in Engine-Assessed Equal Chess Positions},
  pdfauthor={Jesung Park},
}

\begin{document}
\maketitle

\begin{abstract}
\noindent
Among chess opening positions that a strong engine judges essentially equal
(Stockfish~18 evaluation within 10 centipawns of zero, depth-stable) and that
humans actually reach on Lichess (October~2025; 1{,}661 positions, 16.1M
occurrences), human results are not balanced. Positions carry outcome
skews, each the gap between its games' actual results and what the players' ratings
predict, whose directions are stable properties of the naturally-reached position: some positions favour
White, others Black. These skews reproduce
across three re-partitions --- disjoint player-account sets (primary), time,
and disjoint rating bands --- and on an
out-of-sample month eight months later.
On the primary split, each position's skew is measured
once in each account group, and the \emph{replication slope} asks how
well one measurement predicts the other after removing rating and opening-family
effects: one means undiminished carry-over; zero, no linear relation. We find
\textbf{0.69} (family-clustered 95\% CI $[0.65, 0.74]$),
rising to 0.94 on the most-popular, best-measured positions. The slope's value
depends on the position mix. \emph{Existence} is the invariant claim: it survives every tighter
evaluation band, search depth, calibration, and popularity cutoff we test, and
replicates within blitz and rapid separately. The typical skew is small
(median $|\delta| \approx 0.018$, about two percentage points
of White score), yet it reproduces, position by position, across
disjoint accounts. At these positions the disfavoured side also thinks
longer. Even where the evaluation is most confident, it is not a sufficient
statistic for human outcomes. The result is observational, and the causal question is left to a
pre-registered randomised companion study.
\end{abstract}


\section{Introduction}
A modern chess engine reduces a position to a single number, the centipawn
evaluation, and many opening positions evaluate to essentially zero: the
engine calls them dead even. A natural reading is that such positions are
\emph{balanced}: that with two equally matched humans, each side should score about
50\%. That reading conflates two different things. The engine's score describes play
at the engine's level, a level that first surpassed the strongest human players
nearly three decades ago \citep{campbell2002deepblue}. A human outcome is a property of \emph{playable}
difficulty, which need not be symmetric just because the evaluation is. The engine's own
win/draw/loss model already shows that one centipawn value maps to different
empirical W/D/L distributions depending on the material on the board
\citep{stockfish_wdl}: the scalar evaluation is not a sufficient statistic even for
the engine's own outcomes, let alone for humans. Chess merely offers the
cleanest measurement of a general situation: an algorithmic score standing
in as ground truth for a human decision problem.

This paper asks the resulting question directly: when the engine calls an opening
position dead even, do human games from it actually come out even? Take the positions
a strong engine assesses as equal at high search depth (a set we call \sfzero{}),
restrict to those humans actually reach in sufficient numbers, and measure whether
human results from those positions favour one side. They do (as any user of
Lichess's opening explorer can see in the raw win rates), but a measured skew
could still be one of five mundane things rather than a fact about the position as
humans reach it:
sampling noise; a rating artefact (the players on one side of a position happening to
be slightly stronger); a quirk of the particular accounts that reach the position;
a family-level opening effect in disguise (some openings score better for one side, and
every position inside such an opening inherits that family-wide tilt without having
any tilt of its own); or a time-control artefact (a position played mostly at fast
clocks inheriting a tilt that belongs to time pressure, not to the position). Only if all five are ruled out does the skew stand as a real,
reproducible fact about human play from the position.

We rule out the first four with a single, falsifiable reproducibility test, and
the fifth by re-running that test inside each time control separately. For each \sfzero{}
position we split its games into two halves drawn from completely disjoint sets
of player accounts (a separate calendar split of the month provides the temporal
replication axis), and in each half we measure the
position's skew: how far its games' outcomes deviate from what the players' ratings
alone would predict. We then ask whether the first-half skew predicts the second-half
skew. A slope near zero would mean the apparent skew is noise. A positive slope means
the skews measured from two independent groups of players agree, position by position. The test runs
within each opening family (each position is compared only against other
positions from the same opening), so a positive result cannot be a family-level
opening effect. The headline replication slope is \textbf{0.69}, and chance does not plausibly
produce it: a permutation test shuffling only within opening families gives
$p = \textbf{0.001}$, the smallest value it can resolve. The skews
themselves are small, typically about two percentage points of White score.
The finding is their reproducibility across disjoint player groups, not
their magnitude. The result replicates across
the three re-partitions we test (by player, by time, and across disjoint
rating bands) and at both opening-family granularities of the estimator. Closing the fifth explanation, it also replicates separately
inside each time control (Section~\ref{sec:robustness}).

What we defend as the finding is \emph{existence}: the skew remains highly
significant no matter how we perturb the position set or the analysis design,
including at every popularity cutoff, and the
entire per-position pattern reproduces on an out-of-sample month eight months
later, even among accounts absent from
the October analysis entirely (Section~\ref{sec:robustness}). The particular
value 0.69, by contrast, depends on the mix of positions studied: restricting to
more-popular positions (whose skews are measured from more games, with less noise
and therefore less slope-flattening) raises it towards 0.94, so we treat the
replication slope as a mix-specific summary, not a universal constant.

The result is observational and non-causal. We show the skew is real,
rating-adjusted, account-disjoint, and not a family-level opening artefact.
Throughout, ``account-disjoint'' means exactly that: the skew replicates across
disjoint sets of player accounts (one human may hold several), not
that the result is independent of who selects into these positions
(Section~\ref{sec:scope}). We do not
claim the position \emph{causes} it: players who choose to enter a particular
line may be systematically stronger-than-their-rating in that structure (a
repertoire-selection confound that no observational design on naturally-occurring
games can rule out). Establishing causation would require randomly assigning players
to play both sides of a position, the subject of a pre-registered causal companion
study. What we measure is a property of the \emph{naturally-reached position} (the
position together with the players who choose to reach it), not of the position in
isolation.

\paragraph{Contributions.}
\begin{enumerate}[topsep=2pt,itemsep=1pt,leftmargin=1.4em]
  \item A large-scale observational demonstration (26.4M games and 16.1M
    analysed occurrences in the study month, plus 25.0M games and 15.3M
    occurrences in an out-of-sample month) that engine-equal positions carry
    a \emph{reproducible}, rating-adjusted, account-disjoint human outcome
    skew, established by replication across three data re-partitions, two
    family granularities, and an out-of-sample month, rather than by a
    single significance test.
  \item A behavioural counterpart on the clock: at the same engine-equal
    positions, the disfavoured side pays a one-sided, replicated think-time
    cost, measured from the games' per-move clock records.
  \item A measurement design that removes the rating-gap confound by calibration,
    the family-level opening/repertoire confound by comparing positions only
    within their opening family (the sub-family residual is the stated scope
    boundary), the prolific-player confound by a
    player-disjoint split plus a bootstrap over accounts, and the time-control
    artefact by replicating within each time control.
  \item A set of robustness checks separating the invariant claim
    (existence) from the mix-specific one (the replication slope's value):
    membership, calibration, data-quality, dependence, popularity, and engine-family
    checks.
  \item A released, deep-verified \sfzero{} position set with per-position skew
    estimates and complete reproduction provenance (engine build, parameters,
    content fingerprints).
\end{enumerate}

\section{Related work}
The intuition that ``engine-equal'' need not mean ``humanly equal'' has substantial
but scattered precedent.

\paragraph{Decision complexity at fixed structure.}
\citet{barthelemy2025chess960} studies decision complexity across Chess960 starting
positions and reports that engine evaluation and a cognitive-burden measure are only
weakly correlated, with an asymmetry between the two sides. This is the closest
published conceptual precedent, but it differs where it
matters: it concerns opening starting positions rather than
naturally-reached middlegame-entry positions, uses shallow optimal play,
and its side asymmetry is a computed cognitive-load asymmetry, validated
against human thinking times but no game outcomes. It characterises
positions, not results. We supply what it lacks: a measured, replicated human-outcome skew.

\paragraph{Objective difficulty of solved positions.}
\citet{anderson2016assessing,anderson2017assessing} provide the canonical
``objectively known but practically hard'' operationalisation, using tablebases as
ground truth and decomposing human error into skill, time, and inherent difficulty,
finding inherent difficulty dominates. They condition on \emph{solved}
(win/draw/loss-decided) positions and measure move error / blunder rate, not a
directional outcome skew at \eval$\,\approx 0$.

\paragraph{Asymmetry near equality.}
\citet{regan2011intrinsic} give the deepest prior treatment of asymmetry near
equality: raw move error is higher for the side that stands worse, an effect
that scales with the size of the disadvantage. That mechanism is a function of the
evaluation itself, so at \eval$\,\approx 0$ (where neither side stands worse) it
predicts no asymmetry, and it is uniform across positions rather than
position-specific. Our finding sits exactly where their mechanism predicts no effect: at
\eval$\,\approx 0$, game outcomes skew towards a side that varies by position
(with both directions well represented), a position-specific structure their
error--disadvantage relationship cannot generate. Theirs is also a move-error
finding. Ours is a game-outcome finding.

\paragraph{Human move-prediction models.}
Maia \citep{mcilroyyoung2020maia}, Maia-2 \citep{tang2024maia2}, Maia-3
\citep{monroe2026chessformer}, and Allie
\citep{zhang2025allie} model human moves (and, for Allie, thinking time), with
per-rating or skill-aware conditioning. Maia includes a position-level collective
blunder-prediction task, the closest position-level human-difficulty precedent.
Relatedly, \citet{hristova2014difficulty} assess the difficulty of chess tactical
problems for humans. These predict behaviour or rate difficulty. None ranks positions by
a signed human-result asymmetry at fixed evaluation. We use no human move-prediction
model: the skew is established directly by replication across disjoint player groups,
so the result does not depend on any model of human play.

\paragraph{Chess traces as behavioural data.}
Large chess corpora are now an established instrument for behavioural questions
that are hard to study elsewhere, precisely because they pair millions of
naturally-occurring decisions with an objective quality benchmark:
\citet{strittmatter2020lifecycle} recover life-cycle patterns of cognitive
performance from a century of tournament games, and \citet{kunn2023airquality}
identify the effect of indoor air quality on decision quality from moves played
under monitored particulate concentrations. \citet{blasius2009zipf} showed the popularity distribution of opening lines
follows Zipf's law in curated master-game databases, and, at the platform
scale we use, \citet{chowdhary2023quantifying} quantify individual
performance dynamics across more than 120M Lichess games. We use the same trace-plus-benchmark
structure, but ask a question about the positions rather than the players:
whether a property of the decision problem itself survives replication across
disjoint populations.

\paragraph{Centipawn-to-outcome maps and why we calibrate.}
The Lichess Win\% model \citep{lichess_accuracy} and the Stockfish WDL model
\citep{stockfish_wdl} map centipawns to win probability, but neither is
rating-stratified: the former is calibrated on a single 2300-rated benchmark
cohort, the latter on engine self-play. We therefore build an
empirical, per-rating expected-score model rather than inheriting one. Two further
strands support that choice and our colour-aware baseline: the established White
first-move advantage (an empirical White score of $\sim$54\%, $\sim$+35 Elo
equivalent; \citealp{sonas2002firstmove}, a practitioner
analysis, with the advantage also documented at the elite level in
peer-reviewed work, \citealp{gonzalezdiaz2016cognitive})
means the no-skew reference is not 50\%, and the
rating-dependence of conversion supports per-rating calibration.

\paragraph{Engine-derived difficulty measures.}
\citet{barthelemy2025pventropy} reports that move-finding difficulty
(principal-variation entropy) concentrates near $E\!=\!0$ for expert players, and
\citet{barthelemy2025fragility} introduces a position-fragility measure peaking around
move~15. Both quantify how hard a position is to play --- an unsigned notion,
symmetric between the sides (and, for PV entropy, expert-specific and measured over a
much wider $|E|<100$\,cp band than our $|\eval|\le 10$\,cp) --- so neither measures the
signed, directional outcome skew studied here.

\paragraph{The confound we cannot rule out observationally.}
Practitioner analyses of opening statistics have long noted that aggregate
opening win-rates are contaminated by who chooses the opening, the
repertoire-selection bias \citep{yang2018skillbias, d2d4_openingstats}: conditioning on
a line also selects a subset of players, so an opening's aggregate win-rate reflects who
plays it, not its objective merit. Our within-family fixed effects remove the
family-level version of this bias. The residual sub-family version is the
boundary that makes our claim observational (Section~\ref{sec:scope}).

\section{Methods}\label{sec:methods}

\subsection{Data}\label{sec:data}
We use the Lichess standard-rated games dump for \textbf{October 2025}.
We accept games with both players Elo~$\ge 1000$; time control (TC) in
\{blitz, rapid, classical\}; opening positions at $\le 40$ plies; no dedupe
(one occurrence row per reached position per game); ECO-tagged. Sampling is
stride-2, applied upstream of ingest, so the read count below is post-stride.
The funnel:
\textbf{$\sim$91.5M games in the dump $\rightarrow$ 45.8M read (every second game)
$\rightarrow$ 26.4M accepted $\rightarrow$ $\sim$993M
occurrence rows} ($\sim$300M unique positions).

\paragraph{Discovery / confirmation split.}
We divide the month into two calendar halves: \textbf{discovery} (October 1--15)
and \textbf{confirmation} (October 16--31). These names refer only to the
calendar halves, which serve position selection and the temporal replication
axis. The primary replication axis instead splits by account into groups A and B
(Appendix~\ref{sec:splits}). All position selection reads the discovery
half only, so the confirmation half remains a clean replication set.

\paragraph{External month (June 2026).}
A second dump, June 2026, is ingested under the identical inclusion rule and
stride (25.0M accepted games) for the out-of-sample replication of
Section~\ref{sec:robustness}. It is the only external month analysed.

\subsection{Identifying the \sfzero{} set}
The study needs positions that meet two demands at once: the engine must be
confident they are equal, and humans must reach them often enough for a
per-position skew to be measurable. Three steps enforce this. Every
membership evaluation is by Stockfish~18, single-threaded:
\begin{enumerate}[topsep=2pt,itemsep=1pt,leftmargin=1.4em]
  \item \textbf{Selection (discovery-only).} Positions with discovery-half occurrence
    count $\ge 50$: \textbf{253{,}691 candidates}.
  \item \textbf{Shallow screen.} Depth~12, MultiPV~1, $|\eval| \le 50$\,cp:
    \textbf{140{,}202 pass}.
  \item \textbf{Deep verify (decides \sfzero{} membership).} Run on the 7{,}465
    shallow-screen passers with discovery count $\ge 1000$. Membership depends only
    on the engine evaluation: a position is a member when its deepest
    White-point-of-view evaluation lies within 10\,cp of zero and that
    evaluation is stable: the search is deepened along the ladder
    $(12,16,20,25,28)$ and the last three depths must agree within 50\,cp (the mixed
    odd/even ladder guards against the depth-parity oscillation of engine scores).
    The depth-28 evaluation decides membership (MultiPV~4, storing the top four
    moves' evaluations at each depth). Section~\ref{sec:robustness} re-evaluates
    the panel at depths up to 36.
    Result: \textbf{1{,}661 deep-confirmed \sfzero{} positions} (22.3\% of
    7{,}465), the \emph{panel} analysed throughout this
    paper. Empirically, every member also shows
    confirmation-half occurrence $\ge 1000$, a consequence of position
    popularity rather than a selection criterion.
\end{enumerate}
Membership is decided by Stockfish alone, from a pinned build whose identity
is verified in Section~\ref{sec:robustness}. The panel is also re-scored by a
second engine family (lc0) at a fixed node budget
(Section~\ref{sec:robustness}).

\subsection{The occurrence dataset}\label{sec:occurrences}
The unit of analysis is the \emph{occurrence}: one member position reached in
one game. For every occurrence we record the two account ids (casefolded), both
ratings, the time control, the date, the result, the per-move clocks, and the
ECO opening code. October yields \textbf{16{,}135{,}028 occurrences},
\textbf{16{,}134{,}933} of them usable (95 lack a White score: unfinished or
unrecorded results). \textbf{8.99M games} reach at least one member position,
played by \textbf{1{,}144{,}721} distinct accounts (968{,}513 of them in the
games retained by the primary split, Appendix~\ref{sec:splits}). A game can
reach more than one member position (mean 1.80 among contributing games; games
reaching two or more carry 70\% of occurrences), so occurrences from the same
game share its result. Every position's recovered per-half
occurrence count matches an independent tally exactly, and the 95 unusable rows
are spread evenly across time controls, rating buckets, and calendar halves.

\subsection{Expected-score calibration}\label{sec:calib}
A position's raw White score confounds the position with the players who reach
it: if White is, on average, the stronger side in some line, White will score
well there for reasons that have nothing to do with the position. Each game is
therefore scored against a rating-based expectation,
\[
  \mathbb{E}[\text{White score}] = f(\text{rating gap},\ \text{rating level},\ \text{time control}),
\]
a fractional-logit GLM \citep{papke1996fractional} flexible in the rating gap
(a natural cubic spline), the rating level, the time control, and their
interactions (full specification in Appendix~\ref{app:calib}). The two axes that
transfer a calibration across cells fit it on the first cell only and
apply it, unchanged, to the second: the primary axis on group A's 4.04M
usable occurrences, the temporal axis on the discovery half's 7.70M. The
rating-band axis fits no calibration of its own: bands partition the residuals
of the group-A fit, whose fitting sample spans all three bands almost equally.
An empirical curve is fitted rather than the textbook Elo logistic
\citep{elo1978rating} because the latter is misspecified here on three counts:
Lichess ratings are Glicko-2 \citep{glickman2022glicko2}; the empirical White
score at rating parity is $0.515$, not $0.5$, and shifts with time control; and
the gap-to-score relationship moves with rating level and time control.
Held-out predictions track observed White scores closely across
rating-gap$\,\times\,$TC bins, with a parity White edge of 0.515 / 0.510 / 0.505
(blitz/rapid/classical), and the fitted model's transfer to its held-out cell
is checked by that cell's mean residual, reported with the results.

The per-game position residual is
$r_g = \text{realised}_g - \widehat{\mathbb{E}}[\cdot]$ and the position skew is
$\delta_i = \operatorname{mean}_g(r_g)$, always \emph{White-point-of-view} (never
flipped to side-to-move, never symmetrised). The model omits an explicit colour
term (the parity edge is absorbed into the intercept), so $\delta_i$ is a
position's deviation from the average \sfzero{} White score at a given
rating gap. Position identity is not a calibration covariate, so the
calibration cannot absorb position-level skew. It removes the rating-gap
confound, not selection over who plays a position
(Section~\ref{sec:scope}).

\subsection{The replication slope (primary existence test)}\label{sec:gradient}
The primary test asks the question: does a position's skew measured in one cell of a
two-cell split predict its skew measured in the other cell, once opening-family
effects are removed? On the primary axis the two cells are the disjoint
account groups A and B (Appendix~\ref{sec:splits}). The temporal axis
re-runs the identical estimator on the calendar halves. A slope of zero would
mean no linear relation between the two measurements (the apparent skew
would be noise). A slope near one would mean the skew measured in one group of players
carries over at full strength to a disjoint group.

Throughout, a position's \emph{opening family} is the modal ECO code among its
games in the analysis's reference cell (group A on the primary axis), so a
position reached by several move orders sits in a single family. Positions
that are the only member of their family carry no within-family information
and are dropped from the fit (31 of the 1{,}661 on the primary axis, leaving
1{,}630 positions in 118 families; the count shifts by a position or two
across axes because the modal tag is cell-specific, Appendix~\ref{sec:splits}).

The estimator is a \emph{within-family weighted fixed-effects slope}, built in three
steps. First, within each ECO family, subtract the family's weighted mean skew from
every position's $\delta^{A}$ and $\delta^{B}$, so anything that is merely ``this
opening is good for White'', including family-level repertoire selection, drops
out. Second, regress the demeaned group-B skew on the demeaned group-A skew
(with no intercept: demeaning has already centred both variables). Each position is
weighted by its effective sample size: a Kish-style count \citep{kish1965survey} that discounts repeated
games by the same account, taking the \emph{smaller} of the two colours' counts
($n_{\text{eff}} = \min(\text{Kish}_{\text{white}}, \text{Kish}_{\text{black}})$).
This conservative choice keeps a position whose games concentrate on a
few prolific accounts from carrying full weight (formulas in Appendix~\ref{app:neff}). Third, quote standard errors
clustered by opening family \citep{cameron2015clusterguide}, since positions in the same family share context.

The demeaning is conservative. Players choose openings, so a
family's aggregate tilt inseparably mixes family-wide practical difficulty with
repertoire selection. Subtracting it concedes any genuinely family-wide
difficulty to the selection confound, and the claim rests only on the
within-family residual, which is where most of the signal lives
(Section~\ref{sec:results}).

In symbols, writing $\tilde{\delta}_i = \delta_i - \bar{\delta}_{f(i)}$ for the
family-demeaned skew (with $\bar{\delta}_{f(i)}$ the \emph{weighted} mean skew of
position $i$'s opening family $f(i)$, as in the first step),
\[
  \tilde{\delta}^{B}_{i} \;=\; \beta\,\tilde{\delta}^{A}_{i} + \varepsilon_i ,
\]
fitted by weighted least squares with weights $n_{\text{eff},i}$ and family-clustered
standard errors. We call $\beta$ the \emph{replication slope}. It is the paper's
primary estimand, and on the full panel $\hat{\beta} = 0.691$.

Three companions are reported alongside $\hat{\beta}$ on each axis. The slope
with no family demeaning, whose gap from the within-family slope shows how
much of the skew is family-level; the fraction of the anchor cell's skew
variance lying within opening family (an $n_{\text{eff}}$-weighted
decomposition of the demeaned against the total variance); and, because both
cells' skews are estimates and sampling noise attenuates a regression slope
towards zero (regression dilution), the split-half reliability of the
anchor-cell skews --- the estimated fraction of measured skew variance that is
signal rather than sampling noise (ICC; construction in
Appendix~\ref{app:neff}) --- with the disattenuated slope $\hat{\beta}/\text{ICC}$
\citep{spearman1904attenuation,frost2000dilution}. Disattenuation identifies
the presence of a stable component, not its source, and the realised slope
is the estimate of record.

Inference is by permutation: each position's group-B measurement (the skew
$\delta^{B}$ together with its effective sample size) is shuffled among the
positions of its opening family, and the slope is recomputed (re-demeaning and
re-weighting inside each of 1{,}000 draws). The within-family shuffle is the
essential choice: a global or player-level shuffle would destroy family
structure too, and so could not distinguish position-level skew from
opening-family selection. Permutation $p$-values use the add-one correction
$p = (b+1)/1{,}001$ throughout, so every reported $p = 0.001$ means no permuted
draw reached the observed value. Where analytic cluster-robust $p$-values are
quoted they are capped at $10^{-15}$ (the approximation carries no
information at more extreme magnitudes). Inference throughout rests on the
permutation test. Positions differ by two orders of magnitude in
effective sample size, so the test was itself checked: on synthetic data with no
position effect but the observed family structure and per-position precisions,
it rejects at 0.047 / 0.013 against nominal 0.05 / 0.01, so the test is
calibrated. A studentised variant reproduces the headline ($\hat{\beta} = 0.689$ vs.\ 0.691;
construction in Appendix~\ref{app:permcal}).

\subsection{The replication axes}
The effect is required to survive four analyses (three re-partitions of the
data and one re-run of the estimator at a different family granularity), each
ruling out a different alternative explanation:
\begin{itemize}[topsep=2pt,itemsep=1pt,leftmargin=1.4em]
  \item \textbf{Player-disjoint} (primary): could the skew be carried by
    particular accounts? Account ids are randomly assigned to two groups, A
    and B, and a game is kept only when both its players fall in the same
    group, so the two measurements share no accounts at all (49.97\% of usable
    occurrences are retained; construction details in Appendix~\ref{sec:splits}).
  \item \textbf{Opening-family granularity}: could it be repertoire choice?
    Unlike the other three, this is not a data re-partition but an estimator
    check: the within-family estimator of Section~\ref{sec:gradient} is run at
    two granularities, ECO code (primary) and the coarser ECO letter. The
    finer the family definition, the more repertoire selection it removes,
    making ECO code the more conservative run.
  \item \textbf{Temporal} (secondary): does it persist across time? Earlier
    vs.\ later games, using the calendar halves of Section~\ref{sec:data}.
  \item \textbf{Cross-rating-band}: do players of different strengths
    show the same per-position skews? Games are partitioned into three
    disjoint bands by mean rating (L $<$ 1500, M 1500--1800, H $\ge$ 1800,
    set from the rating histogram)
    and the skew is measured independently in each band, with sample-size
    floors (100 games per band, 50 per cell; too-thin cells are excluded
    rather than estimated). Band agreement is summarised by Spearman rank
    correlation, sign agreement, and the within-family cross-band slope. Two
    guard analyses cross the bands with the account split (so the two
    measurements share no games and no accounts) and re-run the primary
    group-A$\to$group-B regression within each band separately
    (Section~\ref{sec:results}).
\end{itemize}
None of the four is an independent dataset: three re-partition a single month
and the fourth re-runs the estimator. The out-of-sample counterpart is the
external-month replication on June 2026 (Table~\ref{tab:headline}). What each
analysis does and does not add is weighed in the Discussion
(Section~\ref{sec:discussion}).

\subsection{Robustness protocol and secondary analyses}\label{sec:protocol}
Every robustness check (Section~\ref{sec:robustness}) re-runs the unchanged
estimator of Section~\ref{sec:gradient} under a perturbation: a membership
or occurrence subset, an added covariate, or an independent re-measurement.
The skew $\delta$ is never reweighted, re-thresholded, or sign-oriented by any
engine quantity. Three constructions recur across the checks and are fixed here.
First, the \emph{replicated high-skew sets}: the positions whose skew reaches
$|\delta| \ge 0.05$ (respectively $0.10$) with the same sign in both cells of
the primary split. These are the practical-significance thresholds tracked
throughout (71 and 4 positions, Section~\ref{sec:results}). Second,
\emph{precision corroboration}: the analytic family-clustered standard error
is checked against two bootstraps (one resampling whole accounts with
Poisson multiplicity, one resampling games), and an influence check re-fits
the slope with each opening family left out in turn (Table~\ref{tab:robust}).
Third, a \emph{covariate-matched placebo}: pseudo-positions that reproduce
each real position's occurrence count and covariate profile while destroying
its chess identity bound what calibration misspecification alone could
manufacture (construction in Appendix~\ref{app:placebo}).

Three further analyses are secondary, specified after the
primary outcome results were known and reported as such: a within-account
re-estimate of the replication slope, demeaning every game outcome by the
account's own mean calibration residual (construction in
Appendix~\ref{app:withinplayer}); a clock-burden
analysis asking whether the side to move spends more clock time when it is
the position's disfavoured side, built from the recorded per-move clocks
(Section~\ref{sec:occurrences}; think-time standardisation, retention, and
interpretable units in Appendix~\ref{app:clock}); and a comparison of the
replicated high-skew set against the remaining members on six pre-specified
engine-derived sharpness measures (definitions and test in
Appendix~\ref{app:sharpness}).

\paragraph{Use of large language models.}
Large-language-model assistants were used, under the author's direction and review, to
help develop analysis code and draft manuscript text. All analyses, numerical results,
and claims were specified and verified by the author, and no large language model is an
author of this work.

\section{Results}\label{sec:results}
Table~\ref{tab:headline} reports the headline test on each axis. On the primary
player-disjoint axis (ECO-code families), one set of players' results predicts a
completely disjoint set of players' results position by position: the replication
slope $\hat{\beta}$ is \textbf{0.691} (family-clustered 95\% CI $[0.646, 0.736]$), and no permuted
draw reaches it ($p = 0.001$). Measurement noise can explain why $\hat{\beta}$ sits below
one: the split-half reliability is ICC $= 0.655$, and correcting for it
(Section~\ref{sec:gradient}) puts the noise-free slope near 1.05, the
95\% interval scaling to $[0.99, 1.12]$. The scaled interval carries only the
slope's sampling noise: the reliability in the divisor is itself an estimate,
treated as known (Appendix~\ref{app:neff}), so the corrected value is read as
``near one'' rather than as a point.
The same test replicates at the coarser ECO-letter family definition (0.771),
on the temporal axis (0.811), and on a different month entirely: June 2026,
eight months out of sample, at 0.904 (details in Section~\ref{sec:robustness}).

\begin{table}[t]
\centering
\footnotesize
\setlength{\tabcolsep}{4.5pt}
\caption{Headline replication slope $\hat{\beta}$ (within-family weighted-FE) across the split axes
and, in the last row, the out-of-sample external month (June 2026 skews
regressed on October's; details in
Section~\ref{sec:robustness}). Family-level opening/repertoire skew is removed
by family fixed effects. ``var.\ frac.'' is the fraction of the anchor cell's
skew variance lying within opening family (group A; full October for the
external month).
\textsuperscript{$\dagger$}\,The ECO-letter definition has only five family
clusters, too few for a credible cluster-robust interval
\citep{cameron2015clusterguide}, so we report its permutation $p$ only.}
\label{tab:headline}
\begin{tabular}{l c c c c c c}
\toprule
Axis (family) & $\hat{\beta}$ & 95\% CI & perm.\ $p$ & ICC & disatten. & var.\ frac. \\
\midrule
\textbf{Primary: player-disjoint (ECO code)} & \textbf{0.691} & $[0.646, 0.736]$ & \textbf{0.001} & 0.655 & 1.05 & 0.62 \\
Player-disjoint, coarser family (ECO letter) & 0.771 & ---\textsuperscript{$\dagger$} & 0.001 & 0.754 & 1.02 & 0.89 \\
Secondary: temporal (ECO code)               & 0.811 & $[0.776, 0.846]$ & 0.001 & 0.796 & 1.02 & 0.58 \\
External: cross-month, Oct$\,\rightarrow\,$June (ECO code) & 0.904 & $[0.872, 0.936]$ & 0.001 & 0.888 & 1.02 & 0.56 \\
\bottomrule
\end{tabular}
\end{table}

\paragraph{The skew is not a family-level opening artefact.}
If the skew were nothing more than ``some openings favour White,'' removing
opening-family means would eliminate it. Instead the replication slope moves only modestly: the raw slope (no family
demeaning) is 0.808 (CI $[0.762, 0.853]$) against the
within-family 0.691 ($n = 1{,}630$
positions in 118 families). Both carry analytic $p < 10^{-15}$ (the cap of
Section~\ref{sec:gradient}). Most of the signal lives inside families:
\textbf{62\%} of the group-A-skew variance is within opening family on the primary
axis (89\% at the ECO-letter granularity; 58\% on the temporal axis). And no single
opening drives the result: leaving out any one family moves the replication slope only within
$[0.682, 0.699]$.

\begin{figure}[!tbp]
  \centering
  \begin{minipage}{0.48\linewidth}
      \centering
      \resizebox{\linewidth}{!}{%
      \begin{tikzpicture}
        \begin{axis}[
            width=8cm, height=8cm,
            title={(a) Observed: disjoint account groups},
            title style={font=\scriptsize},
            xlabel={Group-A skew, family-demeaned $\tilde{\delta}_A$},
            ylabel={Group-B skew, family-demeaned $\tilde{\delta}_B$},
            xmin=-0.175, xmax=0.175,
            ymin=-0.175, ymax=0.175,
            xtick distance=0.05,
            ytick distance=0.05,
            xticklabel style={/pgf/number format/fixed, /pgf/number format/precision=2},
            yticklabel style={/pgf/number format/fixed, /pgf/number format/precision=2},
            axis lines=box,
            axis line style={gray!70},
            tick label style={font=\tiny},
            label style={font=\scriptsize},
            legend style={
              font=\tiny,
              at={(0.97,0.03)},
              anchor=south east,
              draw=gray!60,
              fill=white,
              fill opacity=0.9,
              text opacity=1,
              row sep=1pt,
            },
            legend cell align={left},
            clip=false,
            clip marker paths=true,
            filter discard warning=false,
          ]
          \addplot[
            only marks, mark=*, mark size=0.6pt, color=black!35, opacity=0.55,
            x filter/.expression={\thisrow{class_code}!=2 ? x : nan},
          ] table [x=dA_t, y=dB_t, col sep=comma]
            {figures/scatter_data.csv};
          \addlegendentry{Positions ($n=1624$)}
          \addplot[draw=blue!45!gray, line width=0.4pt, line join=round,
                   fill=blue!9, forget plot]
            table [x=x, y=y, col sep=comma] {figures/scatter_density_p75.csv};
          \addplot[draw=blue!45!gray, line width=0.4pt, line join=round,
                   fill=blue!18, forget plot]
            table [x=x, y=y, col sep=comma] {figures/scatter_density_p50.csv};
          \addplot[draw=blue!45!gray, line width=0.4pt, line join=round,
                   fill=blue!30, forget plot]
            table [x=x, y=y, col sep=comma] {figures/scatter_density_p25.csv};
          \addplot[domain=-0.175:0.175, samples=2, color=gray!45, thin, forget plot]
            ({x},{0});
          \addplot[domain=-0.175:0.175, samples=2, color=gray!45, thin, forget plot]
            ({0},{x});
          \addplot[
            only marks, mark=diamond*, mark size=2.6pt, color=red!75!black,
            x filter/.expression={\thisrow{class_code}==2 ? x : nan},
          ] table [x=dA_t, y=dB_t, col sep=comma]
            {figures/scatter_data.csv};
          \addlegendentry{Showcased, Fig.~\ref{fig:boards} ($n=6$)}
          \addplot[domain=-0.175:0.175, samples=2, color=black!75,
                   line width=0.9pt]
            ({x},{0.690945*x});
          \addlegendentry{Fitted slope $\hat{\beta}$}
          \addlegendimage{area legend, fill=blue!18, draw=blue!45!gray}
          \addlegendentry{KDE core (25/50/75\%)}
          \node[
            font=\scriptsize, align=left, anchor=north west,
            fill=white, fill opacity=0.85, text opacity=1,
            draw=gray!55, inner sep=3pt, rounded corners=1pt,
          ] at (axis cs:-0.170,0.170) {$\hat{\beta} = 0.69$\\
            95\% CI $[0.65, 0.74]$\\permutation $p = 0.001$};
        \end{axis}
      \end{tikzpicture}
      }%
  \end{minipage}\hfill
  \begin{minipage}{0.48\linewidth}
      \centering
      \resizebox{\linewidth}{!}{%
      \begin{tikzpicture}
        \begin{axis}[
            width=8cm, height=8cm,
            title={(b) Shuffled within opening family (one draw)},
            title style={font=\scriptsize},
            xlabel={Group-A skew, family-demeaned $\tilde{\delta}_A$},
            ylabel={Group-B skew, shuffled and demeaned},
            xmin=-0.175, xmax=0.175,
            ymin=-0.175, ymax=0.175,
            xtick distance=0.05,
            ytick distance=0.05,
            xticklabel style={/pgf/number format/fixed, /pgf/number format/precision=2},
            yticklabel style={/pgf/number format/fixed, /pgf/number format/precision=2},
            axis lines=box,
            axis line style={gray!70},
            tick label style={font=\tiny},
            label style={font=\scriptsize},
            clip=false,
            clip marker paths=true,
            filter discard warning=false,
          ]
          \addplot[
            only marks, mark=*, mark size=0.6pt, color=black!35, opacity=0.55,
          ] table [x=dA_t_null, y=dB_t_null, col sep=comma]
            {figures/scatter_data.csv};
          \addplot[draw=blue!45!gray, line width=0.4pt, line join=round,
                   fill=blue!9, forget plot]
            table [x=x, y=y, col sep=comma] {figures/scatter_density_null_p75.csv};
          \addplot[draw=blue!45!gray, line width=0.4pt, line join=round,
                   fill=blue!18, forget plot]
            table [x=x, y=y, col sep=comma] {figures/scatter_density_null_p50.csv};
          \addplot[draw=blue!45!gray, line width=0.4pt, line join=round,
                   fill=blue!30, forget plot]
            table [x=x, y=y, col sep=comma] {figures/scatter_density_null_p25.csv};
          \addplot[domain=-0.175:0.175, samples=2, color=gray!45, thin, forget plot]
            ({x},{0});
          \addplot[domain=-0.175:0.175, samples=2, color=gray!45, thin, forget plot]
            ({0},{x});
          \addplot[domain=-0.175:0.175, samples=2, color=black!75,
                   line width=0.9pt]
            ({x},{0.0169*x});
          \node[
            font=\scriptsize, align=left, anchor=north west,
            fill=white, fill opacity=0.85, text opacity=1,
            draw=gray!55, inner sep=3pt, rounded corners=1pt,
          ] at (axis cs:-0.170,0.170) {$\hat{\beta}$ (this draw) $= 0.02$\\
            no draw of 1{,}000 reaches 0.69};
        \end{axis}
      \end{tikzpicture}
      }%
  \end{minipage}
  \caption{\textbf{The headline test, drawn: one group of players' skews
  predicts a disjoint group's at slope 0.69; shuffling within opening
  family erases the relation.} Both panels plot the estimator's own
  quantities, family-demeaned skews $\tilde{\delta}$
  (Section~\ref{sec:gradient}), for the 1{,}630-position fit sample
  (31 single-member-family positions demean to zero and are excluded).
  (a)~Observed: group-B against group-A skew,
  with the fitted within-family slope $\hat{\beta} = 0.69$ (family-clustered
  95\% CI $[0.65, 0.74]$, permutation $p = 0.001$, mix-specific to this
  panel). Nested contours are kernel-density regions enclosing
  $\approx 25/50/75\%$ of positions; the core stretches along the fitted
  line. Red diamonds mark the six positions showcased in
  Figure~\ref{fig:boards}. (b)~The same construction under one draw of the study's
  permutation null --- $(\delta_B, n_{\text{eff}}^B)$ reassigned within
  each opening family, weights rebuilt, re-demeaned (same KDE bandwidth as
  panel~a): the cloud flattens to slope 0.02, and no draw of 1{,}000
  reaches the observed slope.}
  \label{fig:scatter}
\end{figure}
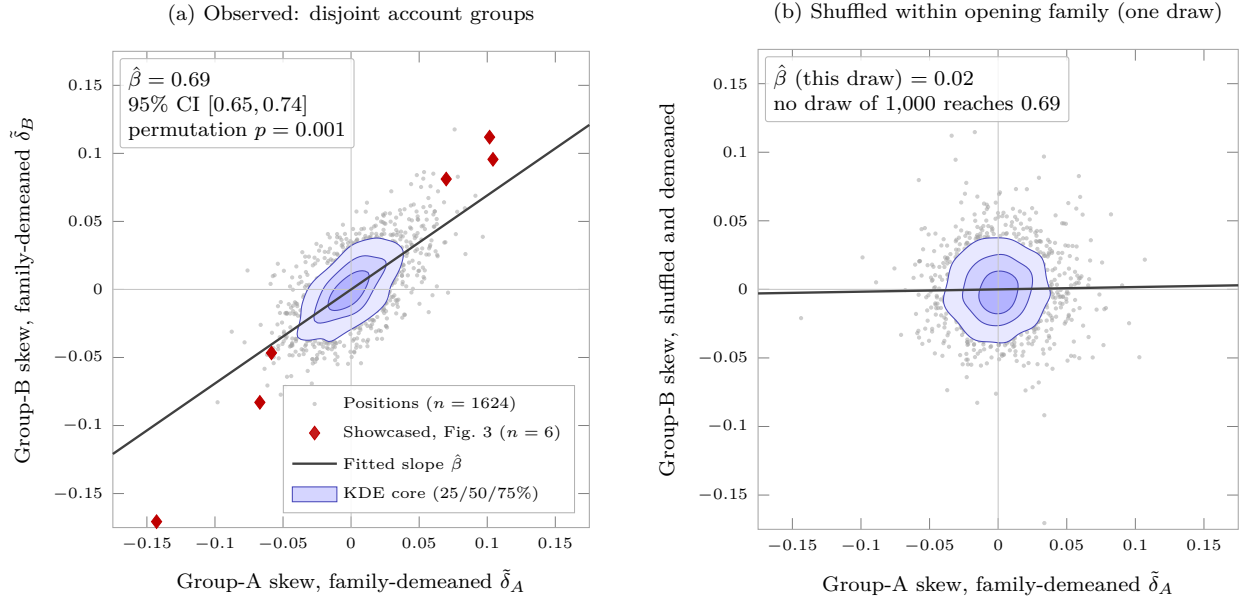

\paragraph{Direction and magnitude.}
Group-B skew rises monotonically across quartiles of group-A skew
(unweighted quartile means $[-0.030, -0.007, +0.006, +0.020]$): the positions that looked worst for White in
one account group look worst in the other, and vice versa. The typical per-position
skew is modest: across the 1{,}661 positions the median $|\delta|$ is
$\approx 0.018$ and the mean $\approx 0.022$ per group, about two percentage points of
White score (the full per-position table is in the released artefact). At the
tail, 71 positions replicate $|\delta| \ge 0.05$ with the same sign in both groups and 4
exceed 0.10. The skew is two-sided rather than a uniform side bias: it favours
White in some positions and Black in others, with both directions well represented
(566 positions tilt towards White in both groups, 671 towards Black). The
mean of $\delta$ is not informative on this point: the calibration's
intercept absorbs any net White edge by construction. The held-out group-B mean
residual ($+0.0002$) is a calibration-transfer check.
Figure~\ref{fig:scatter} draws the headline test. Panel~(a) plots each
position's family-demeaned group-B skew against its family-demeaned group-A
skew (the estimator's own quantities) for the 1{,}630-position fit
sample: the cloud stretches along the fitted line of slope 0.69, with the
six positions showcased in Figure~\ref{fig:boards} marked in red.
Panel~(b) repeats the construction under
one draw of the within-family permutation null of
Section~\ref{sec:gradient}: with the per-position link severed, the cloud
rounds and the fitted slope collapses to 0.02, while no draw of 1{,}000
reaches the observed 0.69.

\paragraph{The skew is rank-stable across rating bands.}
If the skew were an artefact of one player population, players of different strengths
would rank the positions with little relation to one another. Instead the bands agree far
more than that explanation allows. Computing
each position's skew independently within each rating band: Spearman
rank correlation 0.40--0.53 across the three band pairs, sign agreement 0.64--0.68,
cross-band replication slopes 0.38--0.50, every permutation $p = 0.001$.
Even the most distant pair, players below 1500 against players above 1800,
shows rank correlation 0.40 and a slope of 0.38 (per-pair values and intervals
in the released artefact). These slopes sit
below the primary axis for a mechanical reason: in a cross-band comparison
both variables are noisy estimates, and correcting for that reliability puts
the cross-band replication slopes at 0.74--0.94, close to the primary axis. Two further analyses
support this reading. First, a \emph{player-disjoint-clean} variant crosses band with
the account split (e.g.\ band-L skew from group A vs.\ band-M skew
from group B), so the two measurements share no games and no accounts.
All six such combinations stay positive and significant (cross-band replication slopes
0.25--0.36, disattenuated 0.81--1.03). Second, the primary
group-A$\to$group-B regression replicates within each band separately
(within-band replication slopes 0.448/0.409/0.425 for L/M/H, each permutation $p = 0.001$), so
no single stratum carries the effect. Descriptively, mean $|\delta|$ declines only
slightly with rating (0.028/0.025/0.024 for L/M/H). We leave the decline
uninterpreted: the draw rate rises with rating in these data
(3.8\% / 4.2\% / 5.5\% across L/M/H), and that rise compresses measured
skews whatever its source, so a genuine tempering of the skew by stronger
players and a mechanical compression of the outcome scale would look the
same in this comparison.

\paragraph{The skew has a counterpart on the clock.}
At these positions the side to move thinks longer when it sits on the
position's disfavoured side. The measurement behind that sentence uses one
think time per occurrence: the seconds the side to move spends on the single
move it plays \emph{from} the member position, reconstructed from the game's
per-move clock record and increment-corrected (move-level response time is an
established behavioural signal in online chess, \citealp{sigman2010response}).
15.7M of the 16.1M occurrences are usable (97.2\%; the gap is almost entirely
ply-1 occurrences, which have no preceding clock to read --- full accounting
in Appendix~\ref{app:clock}), and, because raw seconds are incomparable
across clock formats, each think is standardised against movers at the same
move number under the same time control, base and increment. A position's
value is the mean over its occurrences. Regressing that mean on the
position's mover-point-of-view skew ($+\delta$ when White is to move,
$-\delta$ when Black is), within opening family and with controls for
$|\delta|$, side to move, ply and mover rating (1{,}629 positions in 118
families), gives $\hat{\beta} = \mathbf{-3.54}$ (family-clustered 95\% CI
$[-4.86, -2.23]$; $p = 0.0005$, the floor of 2{,}000 draws, under both
permutation nulls of Appendix~\ref{app:clock}). The cost is one-sided: the
coefficient on $|\delta|$ is indistinguishable from zero ($-1.07$,
$t = -0.89$), so skewed positions are not slow for both players --- the extra
time sits with the disfavoured side alone. In interpretable units, a 0.10
swing in mover-POV skew corresponds to roughly 23\% more think time. At the
replicated-skew positions, measured against low-$|\delta|$ control positions
from the same opening families, the favoured side moves \emph{faster} than
neutral (2.5 cell-standardised seconds against the controls' 3.2) while the
disfavoured side moves slower (4.2\,s; a 67\% think-time gap) and, on
the subset with at least two family-matched controls, takes long thinks 2.3
times as often (12.0\% against 5.3\%; thresholds in
Appendix~\ref{app:clock}). Both deviations from neutral hold in direction on
every outcome scale. However, which side deviates more is scale-dependent, so
only the gap between the sides is quoted as a magnitude. Three further facts
are established for the reading in Section~\ref{sec:clockscope}: the
coefficient is statistically flat across time controls ($-3.37$ / $-4.06$ /
$-3.18$ for blitz/rapid/classical, and stable across weighting schemes and
family definitions); it is unchanged when the think times and the skew are
computed from disjoint halves of the data ($-3.38$ and $-3.53$); and
demeaning each occurrence by the mover's own average think time across the
member set (leave-one-out) retains 84\% of the coefficient ($-2.99$,
$t = -5.46$).

\subsection{Robustness and sensitivity}\label{sec:robustness}
Each check re-runs the unchanged estimator under the protocol of
Section~\ref{sec:protocol}. Unless stated otherwise the pre-specified pass
rule is a replication slope within $\pm 0.05$ of the
0.691 headline and within-family-block permutation $p \le 0.01$. Five checks
breach the $\pm 0.05$ band (the fast-moving-rating filter, the
one-position-per-game re-fit, the covariate-means re-fit, the
popularity-cutoff grid, and the within-time-control re-fits) and are reported
below as band failures. Any ICC-based reading alongside them
is interpretation rather than a pass. With one exception these checks
are nested subsets and re-fits of one panel, and the
recurring $p = 0.001$ is the resolution floor of 1{,}000 permutation draws
(Section~\ref{sec:gradient}). The exception is the external-month replication
(below).
Table~\ref{tab:robust} summarises the checks. Those whose reading
needs more than a verdict follow in prose. The complete result files and
constructions behind every check are in the archived dataset (see Data and code availability).

\begin{table}[p]
\centering
\footnotesize
\setlength{\tabcolsep}{4pt}
\caption{The robustness checks. Each row re-runs the unchanged estimator of
Section~\ref{sec:gradient} under the stated perturbation; \textsc{pass} = replication
slope within $\pm 0.05$ of the 0.691 headline and within-family permutation
$p \le 0.01$. Throughout, perm.\ $p = 0.001$ means 0 of 1{,}000 draws reached the
observed slope, the test's resolution floor.}
\label{tab:robust}
\begin{tabular}{>{\raggedright\arraybackslash}p{3.9cm} p{6.7cm} c l}
\toprule
Check & Result & perm.\ $p$ & Verdict \\
\midrule
Tighter eval bands\textsuperscript{a} ($\le$5/7/8\,cp; 917/1{,}178/1{,}325 members) &
slope 0.679 / 0.679 / 0.686 & 0.001 & \textsc{pass} \\
Deeper search (depth 36; 267 positions drifting out of band dropped) &
slope 0.666 ($\Delta = 0.025$) & 0.001 & \textsc{pass} \\
Colour-aware calibration refit (adds side-to-move terms;
Appendix~\ref{app:calib}) &
slope 0.684 ($\Delta = 0.007$) & 0.001 & \textsc{pass} \\
Depth-28 eval as covariate &
slope 0.687 ($\Delta = 0.004$); eval coefficient not significant once the
group-A skew is included; within-band eval bounded at 11.1\% of
within-family variance (point share 2.1\%) & 0.001 & \textsc{pass} \\
Provisional-account filter\textsuperscript{b} (either account $<$10 games in the
month) &
slope 0.648 ($\Delta = 0.043$) & 0.001 & \textsc{pass} \\
Fast-moving-rating filter\textsuperscript{b} (Elo range $>$150; removes 57\% of
occurrences) &
slope 0.511 (0.443 with both filters); ICC 0.41 / 0.32; disattenuated $\ge 1$ &
0.001 & \textsc{fail} (band) \\
High-skew sets under BH--FDR\textsuperscript{e} \citep{benjamini1995fdr} &
71/71 at $q < 0.05$ (max $q = 0.0017$) and 4/4; the permutation null expects 24.3 and
1.5 & 0.001 & \textsc{pass} \\
Player-account bootstrap\textsuperscript{c} (400 draws) &
SE 0.0183 vs.\ analytic family-clustered 0.0227 & --- & corroborates \\
Game-cluster bootstrap\textsuperscript{c} (400 draws, 4.49M clusters) &
SE 0.0200 vs.\ 0.0227 & --- & corroborates \\
Second engine family\textsuperscript{d} (lc0 v0.32.1, 800 nodes) &
1{,}660/1{,}661 concordant (99.9\%); high-skew sets 71/71 and 4/4; random
control 200/200 & --- &
\textsc{pass} \\
Engine build identity &
binary SHA-256 matches the official Stockfish~18 release; the depth-36 rerun
reproduces the evaluations at all shared depths exactly & --- & exact \\
External month\textsuperscript{f} (2026-06; panel, pipeline, calibration fixed
in advance) &
cross-month slope 0.904 (Oct.\ cal.) / 0.901 (June refit); all 1{,}661
members recur & 0.001 & \textsc{pass} \\
\bottomrule
\end{tabular}

\medskip
\begin{minipage}{0.97\linewidth}
\raggedright\footnotesize
\textsuperscript{a}\,Wider bands (15--20\,cp) are not re-analysable by design:
positions outside 10\,cp were screened out before the human-outcome join.
\textsuperscript{b}\,Weak proxies for the unobserved Glicko-2 rating reliability. The
volatility filter mechanically removes the most-active accounts (observed range grows
with games played), so its slope decline tracks the ICC decline, a bounded probe
rather than a full discharge.
\textsuperscript{c}\,The account bootstrap resamples whole accounts with
Poisson(1) multiplicity (a draw can drop or duplicate an account outright) over
the 875{,}350 accounts appearing as the mover at an analysed position in a
split-retained game (a subset of the 968{,}513 appearing in either colour).
The game-cluster bootstrap resamples games. Both corroborate the analytic
family-clustered SE, so the reported precision does not rest on a small set of
prolific accounts. SEs are quoted around the realised estimate. However, bootstrap
percentile intervals are invalid here because each resample drops $\sim$37\% of every
position's games and mechanically attenuates the resampled slopes (centre 0.527).
\textsuperscript{d}\,A membership cross-check at a fixed 800-node search budget
with a pre-specified concordance criterion ($|Q| \le 0.15$, lc0's win-probability
scale), not a re-verification at Stockfish-comparable depth.
\textsuperscript{e}\,Per-position $p$: inverse-variance-pooled skew across the two
groups, two-sided $z$; BH step-up over all 1{,}661 positions --- 520 positions
at $q < 0.05$ overall --- with the practical-size sets read off that single run. The ``permutation null expects''
counts come from the within-family characterisation shuffle (1{,}000 draws), not
from BH.
\textsuperscript{f}\,The only check drawing on a different month of games. The
cross-month slope is reported under both the October calibration and a
June refit. The $\pm 0.05$ band does not apply here (see the external-month
paragraph).
\end{minipage}
\end{table}

\paragraph{External-month replication (out-of-sample).}
Every other check in this section re-analyses October 2025, so a month-specific
common cause (a client release, an interface change, a fashion in openings)
would survive them all. The direct test, the final row of
Table~\ref{tab:headline}, re-runs the pipeline on the June 2026 month
of Section~\ref{sec:data}, with the panel, the
membership decisions, and every estimator choice fixed before the June data
were read. Nothing was lost to
opening-fashion drift: all 1{,}661 members recur in June (15.3M occurrences),
and all clear the 50-occurrence floor in both months. The
\emph{cross-month replication slope} --- each
position's June skew regressed on its October skew, within family and weighted
following Section~\ref{sec:gradient} --- is \textbf{0.904} (family-clustered
95\% CI $[0.872, 0.936]$; permutation $p = 0.001$; 1{,}630 positions in 119
families\footnote{119 here against 118 on the primary split: family
assignment is cell-specific (Section~\ref{sec:gradient}), and one position's
modal code shifts between the primary's group-A half and the
full October month (D30 to D32), promoting a dropped singleton family to a
retained two-member one; an offsetting shift (B70 to B72) leaves both samples
at 1{,}630 positions, differing in two members.}) under the October calibration, and 0.901 under a calibration
refit on June, so calibration drift is immaterial: the October calibration's
held-out parity edges in June are 0.515/0.510/0.503 across blitz/rapid/classical,
against 0.515/0.510/0.505 in October. The October skew map
reproduces in games played eight months later: 76 positions replicate
$|\delta| \ge 0.05$ with the same sign in both months, 4 at $\ge 0.10$. Nor is
the replication carried by returning players: restricting June to games in
which neither account appears among the October analysed games (13.7\% of June
occurrences; half of June accounts are new) leaves all 1{,}630 positions
above the occurrence floor and gives a slope of \textbf{1.01} (95\% CI
$[0.95, 1.08]$; permutation $p = 0.001$): the skew map reproduces in players
absent from the October panel entirely. The
full-month slope of 0.904 sits above the 0.691 headline for the reason
Figure~\ref{fig:dose}
predicts: both of its cells are full months, so each skew is measured with roughly
twice the data of the halved account split (reliability 0.888 vs.\ 0.655), and
less measurement noise means less regression dilution (disattenuated
$\approx 1.02$, near one as on the primary axis). The $\pm 0.05$ band of this section is
defined for same-month re-fits of the 0.691 estimand and is not the yardstick
here.

\paragraph{Within-band evaluation.}
A natural worry is that the skew is residual engine evaluation inside the
$\pm 10$\,cp band. The
evaluation is not entirely uninformative there: regressing each position's skew
on its depth-28 evaluation within family gives $5.3\times10^{-4}$ expected-score
points per centipawn (95\% CI $[2.9, 7.7]\times10^{-4}$). But the component is
small: it accounts for 2.1\% of the within-family skew variance, an 8\,cp
within-band difference implies about 0.004 of skew (an order of magnitude below
the 0.05 replication threshold), and controlling for it leaves the slope at 0.687
(Table~\ref{tab:robust}). Within-band evaluation can neither generate the headline
slope nor manufacture the replicating positions.

\paragraph{Covariate-matched placebo.}
Calibration misspecification alone could, in principle, manufacture stable
skews. A position
whose player population sits where the calibration mis-fits would carry a stable
$\delta$ in both groups with no position-level effect. Re-running the unchanged
estimator on \emph{pseudo-positions} that match each real position's covariate
profile but carry no chess identity (construction in Appendix~\ref{app:placebo})
gives a placebo slope of 0.046 against the 0.691 headline. A positive control
that also matches on ply, a near-identifier of position, rebuilds 0.350, so the
placebo has power to detect identity-driven co-skew. A companion re-fit adds
each position's covariate means directly to the fit (covariate list in
Appendix~\ref{app:placebo}): the slope drops to \textbf{0.599}. We read
this band failure as a measurement: at most $\approx$13\% of the
headline slope is collinear with stable covariate composition, an upper bound
on the composition channel left open in Section~\ref{sec:scope}. Misspecification alone cannot produce the replication slope. However, a
composition component of up to that size can coexist with it.

\paragraph{Shared-game dependence.}
Positions reached in the same game share that game's single outcome
(Section~\ref{sec:occurrences}). A game lies entirely within one cell of every split,
so this cannot correlate a position's two measurements, but it could overstate
the effective information behind them. The game-cluster bootstrap (Table~\ref{tab:robust})
shows that game-level dependence does not widen the reported precision, and a
one-position-per-game re-fit removes the shared-outcome duplication entirely:
keeping one random member occurrence per game yields slopes of 0.540--0.576
across five draws (median \textbf{0.564}, outside the $\pm 0.05$ band, each with
family-clustered $p < 10^{-15}$, the analytic floor). The drop is the size expected from discarding
$\approx$44\% of occurrences (the disattenuated slope is unchanged,
$\approx$1.01--1.12), so existence is unaffected on the game axis.

\paragraph{Account-level outcome structure.}
A related question is whether the skew rides on persistent account-level
differences --- accounts that systematically over- or under-perform their
calibration prediction, distributed unevenly across positions. Demeaning
every outcome by the account's own leave-one-out mean residual, both colours
at once (construction in Appendix~\ref{app:withinplayer}), retains 98.4\% of
occurrences. On those identical rows the slope falls from 0.675 (the plain
fit re-run on this retained sample) to
\textbf{0.526} (95\% CI $[0.47, 0.58]$; permutation $p = 0.001$). Demeaning
lowers the slope by roughly a fifth: about a twentieth of the slope is
mechanical attenuation from the construction itself, and the rest of the
drop is an upper bound on
persistent account-level structure, since account means also absorb genuine
position skew through an account's repeated play of particular lines. The
resulting 0.526 sits outside the $\pm 0.05$ band, which is defined for re-fits
of the 0.691 estimand, not for this demeaned variant. The within-account slope
remains decisively nonzero: existence is unaffected on the account axis.

\paragraph{Line nesting and transposition dependence.}
Positions on a shared forced sequence are reached by nearly the same games, so
their residuals are correlated beyond what family clustering absorbs: most
clearly the two Traxler Counterattack positions, consecutive plies of one
forced line and therefore one piece of evidence, not two. Collapsing the
panel to the best-measured position of each line-cluster (games-overlap
threshold 0.50; construction in Appendix~\ref{app:nesting}) leaves
814 positions, on which the
primary slope is \textbf{0.681} (95\% CI $[0.621, 0.741]$; permutation
$p = 0.001$) against 0.691 on the full panel, and moving the overlap
threshold to 0.25 or 0.75 changes it only to 0.687 or
0.669. The replicator counts do shrink: the 71 positions replicating at
$|\delta| \ge 0.05$ occupy 46 distinct line-clusters and the four at
$|\delta| \ge 0.10$ occupy \textbf{three}, so we count three independent strong
replicators rather than four. The slope itself does not depend on the nesting.

\paragraph{Time-control stratification.}
A position$\times$time-control interaction (sharp lines favouring one side
specifically at fast clocks) could mimic a time-control-independent
position property. Re-running the primary fit inside each time control
separately, the skew replicates independently in both populated strata: blitz
(76\% of occurrences) gives \textbf{0.601} (CI $[0.543, 0.658]$) and rapid
(23\%) gives \textbf{0.451} (CI $[0.390, 0.512]$), each at the permutation floor
and both outside the $\pm 0.05$ band. The lower within-stratum slopes reflect
reliability loss, not signal loss: the disattenuated slopes ($\approx 1.08$ and
$\approx 1.23$) return to one or above, as the full-sample value does
($\approx 1.05$, Table~\ref{tab:headline}). Classical
(0.9\% of occurrences; 90 surviving positions) is uninformative (slope 0.20, CI
$[-0.12, 0.53]$, $p = 0.19$) rather than evidence in either direction. Existence
is not carried by a time-control interaction.

\paragraph{Popularity-cutoff sensitivity.}
The panel conditions on discovery-half occurrence $\ge 1000$. Raising that cutoff in
steps, the replication slope rises
monotonically from 0.691 through \textbf{0.746} ($\ge 1500$, $n = 1{,}058$),
\textbf{0.784} ($\ge 2000$, $n = 749$), \textbf{0.854} ($\ge 3000$, $n = 452$) and
\textbf{0.895} ($\ge 5000$, $n = 258$) to \textbf{0.944} ($\ge 8000$, $n = 150$),
each with permutation $p = 0.001$. The intraclass correlation rises in step,
from 0.66 to 0.87 across the same cutoffs, and the disattenuated slope stays
between 1.05 and 1.09 throughout, every interval covering one. This formally fails the $\pm 0.05$ band,
but in the direction opposite a winner's-curse artefact, which would collapse towards zero
or lose significance. Figure~\ref{fig:dose} and Section~\ref{sec:discussion}
give the reading. Existence
holds at every cutoff. However, the 0.691 headline is quoted throughout as specific to the
$\ge 1000$ panel mix.

\section{Discussion}\label{sec:discussion}
\paragraph{What the skew adds to the public record.}
In one sense the raw observation is folklore: Lichess's own opening explorer
reports a win rate for every position on this paper's boards, and club players
know that certain lines score better than they ``should''. The explorer number,
however, describes a position's traffic as much as the position: it pools
players of different strengths, sides of different ratings, and the repertoire
cultures that steer particular players into particular openings. What this
study adds is the accounting the explorer cannot perform. The skew that remains
after rating calibration is not a strength artefact. The 62\% of its variance
lying within opening family is finer-grained than opening choice itself. And
its replication across disjoint account groups, and across a
month eight months out of sample, makes it a stable property of human play
from the position rather than a fluctuation of one population or one period. The
folklore survives an audit it had never been given, and turns out to be
position-level, not opening-level.

\paragraph{Machine evaluation as ground truth.}
For readers outside chess, the interest of that audit is what it says about
algorithmic scores in general. Chess is an extreme case of a now-common
arrangement: a machine value function (here among the most authoritative
machine assessments in any human decision domain) serves as ground truth for
human decision problems, in training tools, matchmaking, difficulty ratings,
and commentary. Our panel consists of the positions where that ground truth
is clearest: the engine calls them level, at evaluations stable
across the depths that decide membership. Human outcomes from exactly these positions carry
reproducible, directional structure that the evaluation does not represent. A
scalar machine assessment, however deep, summarises play at the machine's
level. It is not a sufficient statistic for human results, and the gap is not
noise, because it replicates.

Three results carry the behavioural content of that conclusion. Most of the
skew's variance lies within opening family, so it attaches to specific decision
problems rather than to the openings players choose. The skew has an
at-the-board counterpart: the disfavoured side pays a one-sided cost in
thinking time (Section~\ref{sec:clockscope}), so the imbalance is visible in
the act of playing, not only in accumulated results. And the replicated set is
not a catalogue of known traps: sharpness marks only the panel's high-skew
tail, and
calm positional structures carry replicated skews of comparable size
(Section~\ref{sec:highskew}). What none of this resolves is whether the
property belongs to the position or to the players who select into it: the
boundary drawn in Section~\ref{sec:scope}, and the question the randomised
companion study is designed to answer.

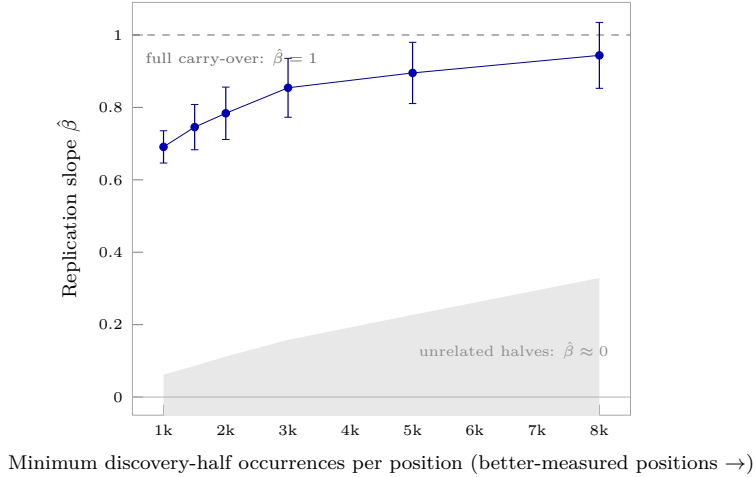
\begin{figure}[!tbp]
  \centering
  \begin{minipage}{0.62\linewidth}
      \centering
      \resizebox{\linewidth}{!}{%
      \begin{tikzpicture}
        \begin{axis}[
            width=8.6cm, height=7.4cm,
            xlabel={Minimum discovery-half occurrences per position (better-measured positions $\rightarrow$)},
            ylabel={Replication slope $\hat{\beta}$},
            xmin=500, xmax=8500,
            ymin=-0.05, ymax=1.09,
            xtick={1000,2000,3000,4000,5000,6000,7000,8000},
            xticklabels={1k,2k,3k,4k,5k,6k,7k,8k},
            ytick={0,0.2,0.4,0.6,0.8,1.0},
            minor tick num=0,
            axis lines=box,
            xtick pos=bottom, ytick pos=left,
            axis line style={gray!70},
            tick label style={font=\tiny},
            label style={font=\scriptsize},
          ]

          \addplot[draw=none, fill=gray!18, forget plot] coordinates {
            (1000,0.0619) (1500,0.0861) (2000,0.1121)
            (3000,0.1580) (5000,0.2271) (8000,0.3288)
            (8000,-0.3160) (5000,-0.2125) (3000,-0.1400)
            (2000,-0.0919) (1500,-0.0719) (1000,-0.0509)
          } --cycle;
          \addplot[domain=500:8500, samples=2, gray!60, thin, forget plot]
            ({x},{0});
          \node[font=\tiny, color=gray!90, anchor=east]
            at (axis cs:8300,0.13) {unrelated halves: $\hat{\beta}\approx 0$};

          \addplot[domain=500:8500, samples=2, dashed, black!60, thin, forget plot]
            ({x},{1.0});
          \node[font=\tiny, color=black!60, anchor=north west]
            at (axis cs:560,0.988) {full carry-over: $\hat{\beta}=1$};

          \addplot[
            color=blue!55!black, thin, mark=none, forget plot,
          ] coordinates {
            (1000,0.690945) (1500,0.745589) (2000,0.783723)
            (3000,0.854176) (5000,0.895172) (8000,0.943732)
          };

          \addplot[
            only marks,
            mark=*,
            mark size=1.5pt,
            color=blue!70!black,
            error bars/.cd,
              y dir=both, y explicit,
            error bar style={color=blue!55!black, line width=0.5pt},
          ] coordinates {
            (1000,0.690945) += (0,0.044565) -= (0,0.044565)
            (1500,0.745589) += (0,0.062556) -= (0,0.062555)
            (2000,0.783723) += (0,0.072382) -= (0,0.072381)
            (3000,0.854176) += (0,0.081386) -= (0,0.081385)
            (5000,0.895172) += (0,0.084571) -= (0,0.084570)
            (8000,0.943732) += (0,0.091017) -= (0,0.091017)
          };

        \end{axis}
      \end{tikzpicture}
      }%
  \end{minipage}
  \caption{\textbf{Measured with less noise, the replication slope approaches one.} Each point recomputes the replication slope $\hat{\beta}$ of Section~\ref{sec:gradient} on the subset of positions whose discovery-half occurrence count meets the cutoff (selection reads discovery-half counts only, preserving the held-out seal); bars are family-clustered 95\% confidence intervals, and the surviving panel shrinks from 1{,}661 positions ($\ge$1{,}000) to 150 ($\ge$8{,}000). $\hat{\beta} = 1$ (dashed) would mean each position's group-A skew carries over at full strength; the shaded band is the within-family permutation null (per-cutoff mean $\pm 2$ s.d., 1{,}000 draws), centred on zero (lower half cropped at the axis floor). The slope rises monotonically from 0.69 to 0.94 (the interval at the strictest cutoff includes one), consistent with regression dilution fading on better-measured positions; a winner's-curse (noise-mining) artefact would collapse towards zero instead.}
  \label{fig:dose}
\end{figure}

\paragraph{Interpretation.}
The realised replication slope of 0.691 shows that a position's skew measured
in one account group carries over, near full strength once measurement noise
is accounted for, to players it shares no accounts with (disattenuated
$\approx 1$). The effect replicates on every axis we test (within
opening family at both granularities,
over time, across the player-disjoint split, and across disjoint rating bands), but
these are not independent confirmations. The player-disjoint and temporal axes
re-partition the same month of games and so corroborate rather than independently
confirm. The cross-rating-band axis is the one drawing its two
measurements from largely non-overlapping player populations (bands are assigned
per game by mean rating, so an account can contribute to more than one band). The
out-of-sample counterpart is the external-month replication of
Section~\ref{sec:robustness}.
Figure~\ref{fig:dose} shows the sharpest version of the mix-specificity pattern:
recomputing $\hat{\beta}$ on progressively better-measured subsets of the panel
raises it monotonically from 0.69 to 0.94 (the interval at the strictest cutoff
includes one) while existence holds unchanged at every cutoff
(Section~\ref{sec:robustness}). The noise-corrected slope confirms the dilution reading:
disattenuated for reliability, the slope is flat near one at every cutoff, so
the rise tracks measurement reliability rather than any popularity-linked
change in the effect. The flatness, not the corrected level, carries this
reading: a popularity-linked effect would have to track the reliability
gradient at every cutoff to mimic it. A winner's-curse or noise-mining artefact
would drift the opposite way, towards zero. The diagnostic separates the skew from
noise artefacts, not from selection (Section~\ref{sec:scope}).

\subsection{The high-skew positions}\label{sec:highskew}
The replicated positions are recognisable opening structures, and they are
not all of one kind. Figure~\ref{fig:boards} shows six of them, chosen from
the 71-position replicated set to span its character (no two sharing a
line-cluster or ECO code). The top row is sharp, tactical territory: the
Englund Gambit Complex (A40), the fork trick of the Italian Four Knights
(C47), and a line of the Nimzo-Larsen Attack (A01). It
includes the panel's largest skews, with human results running 8 to 18
percentage points of expected score from what the players' ratings predict. The bottom
row is quiet: a Giuoco Pianissimo (C50), an Advance Caro-Kann (B12), and a
Colle-type Queen's Pawn setup (D00). These are calm, positional structures with no
tactic in sight, still carrying replicated skews of 8 to 11 points.

These skews can be translated into a currency players know: rating. Near parity, the
fitted calibration curve climbs by roughly 0.1 expected-score points per
hundred rating points,\footnote{0.08 to 0.11 across time-control and
rating-level cells, all flatter than the textbook Elo logistic.} so a
replicated skew of 0.05 (the
threshold all 71 positions clear in both account groups) is the
expected-score shift of a fifty-point rating gap, and the largest replicated
skews correspond to gaps of 150 points or more. The conversion only sizes the skew
in familiar units: the skew is what remains after the curve is
applied, not a rating difference. The engine calls all six level; human
play from them is not.
\newcommand{\headlineboardwidth}{4.2cm}
\newcommand{\lastmovecolor}{orange!30}
\begin{figure}[!tbp]
  \centering
  \setlength{\tabcolsep}{2pt}%
  \begin{tabular}{ccc}
    \begin{minipage}{0.31\linewidth}
      \centering
      \resizebox{\headlineboardwidth}{!}{%
        \chessboard[
          setfen={2kr2nr/ppp1qppp/2nb4/8/4P1b1/2N2N2/PPP1BPPP/R1BQK2R w KQ - 5 8},
          showmover=false,
          color=\lastmovecolor,
          colorbackfields={c8,d8}
        ]}
      \\[2pt]
      \footnotesize (a) A40 --- Englund Gambit Complex\\
      \footnotesize White to move; $\delta_A = -0.150$, $\delta_B = -0.177$
    \end{minipage}
    &
    \begin{minipage}{0.31\linewidth}
      \centering
      \resizebox{\headlineboardwidth}{!}{%
        \chessboard[
          setfen={r1bqkb1r/pppp1ppp/2n5/4p3/2B1N3/5N2/PPPP1PPP/R1BQK2R b KQkq - 0 5},
          showmover=false,
          color=\lastmovecolor,
          colorbackfields={e4}
        ]}
      \\[2pt]
      \footnotesize (b) C47 --- Italian Four Knights, fork trick, after 5.Nxe4\\
      \footnotesize Black to move; $\delta_A = -0.080$, $\delta_B = -0.098$
    \end{minipage}
    &
    \begin{minipage}{0.31\linewidth}
      \centering
      \resizebox{\headlineboardwidth}{!}{%
        \chessboard[
          setfen={r1bqk1nr/ppp2ppp/2nb4/1B1pp3/5P2/1P2P3/PBPP2PP/RN1QK1NR b KQkq - 0 5},
          showmover=false,
          color=\lastmovecolor,
          colorbackfields={f4}
        ]}
      \\[2pt]
      \footnotesize (c) A01 --- Nimzo-Larsen Attack, after 5.f4\\
      \footnotesize Black to move; $\delta_A = +0.105$, $\delta_B = +0.117$
    \end{minipage}
    \\[8pt]
    \begin{minipage}{0.31\linewidth}
      \centering
      \resizebox{\headlineboardwidth}{!}{%
        \chessboard[
          setfen={r1bq1rk1/ppp2ppp/2np1n2/2b1p1B1/2B1P3/2NP1N1P/PPP2PP1/R2QK2R b KQ - 0 7},
          showmover=false,
          color=\lastmovecolor,
          colorbackfields={h3}
        ]}
      \\[2pt]
      \footnotesize (d) C50 --- Giuoco Pianissimo, after 7.h3\\
      \footnotesize Black to move; $\delta_A = +0.108$, $\delta_B = +0.098$
    \end{minipage}
    &
    \begin{minipage}{0.31\linewidth}
      \centering
      \resizebox{\headlineboardwidth}{!}{%
        \chessboard[
          setfen={r2qkbnr/pp3ppp/2n1p3/3pP3/3P2b1/5N2/PP2BPPP/RNBQK2R w KQkq - 0 8},
          showmover=false,
          color=\lastmovecolor,
          colorbackfields={e6}
        ]}
      \\[2pt]
      \footnotesize (e) B12 --- Caro-Kann Advance, after 7...e6\\
      \footnotesize White to move; $\delta_A = -0.101$, $\delta_B = -0.091$
    \end{minipage}
    &
    \begin{minipage}{0.31\linewidth}
      \centering
      \resizebox{\headlineboardwidth}{!}{%
        \chessboard[
          setfen={rnbqkb1r/ppp2ppp/4pn2/3p4/3P4/3BP3/PPPN1PPP/R1BQK1NR b KQkq - 3 4},
          showmover=false,
          color=\lastmovecolor,
          colorbackfields={d2}
        ]}
      \\[2pt]
      \footnotesize (f) D00 --- Colle-type Queen's Pawn, after 4.Nbd2\\
      \footnotesize Black to move; $\delta_A = +0.081$, $\delta_B = +0.096$
    \end{minipage}
  \end{tabular}
  \caption{Six showcased replicating positions ($|\delta| \ge 0.05$ with the
    same sign in both account groups), chosen to span the panel's character:
    the top row is sharp, tactical structures; the bottom row quiet,
    positional ones. No two share a line-cluster or ECO code
    (Appendix~\ref{app:nesting}). White is at the bottom in all panels;
    each label states the side to move, and the shaded square marks the
    destination of the last move played (king and rook squares for castling,
    panel a). $\delta_A$ and $\delta_B$ are the White-point-of-view
    expected-score residuals in the two account groups: panels (a), (b),
    (e) favour Black; (c), (d), (f) favour White. These six are the red
    diamonds of Figure~\ref{fig:scatter}.}
  \label{fig:boards}
\end{figure}

Sharpness does, however, separate the replicated set statistically. The six
pre-specified measures, computed from the stored engine analysis, fall into
two groups: three ask how concentrated the position's resources are in a
single move (the evaluation gap from the engine's first candidate move to
its fourth, the gap from its first to its second, and the entropy of the
engine's preference over its four candidates), and three ask how settled the
search is (the standard deviation and range of the evaluation across the
depth ladder, and its change over the ladder's last rung). Comparing the 71
replicating positions against the remaining members, always within the opening
families containing both classes (definitions, test, and effective sample
in Appendix~\ref{app:sharpness}), three of the
six separate the sets: among replicators the engine's preference
concentrates in a single move (lower candidate entropy; Cliff's
$\delta = -0.31$, $p = 0.008$), and its evaluation settles less from one
depth to the next (Cliff's $\delta = +0.25$ and $+0.23$ for the ladder
standard deviation and range; both $p = 0.001$), a signature of tactical
rather than positional balance. All three survive multiple-testing
correction across the six (Benjamini--Hochberg, $q \le 0.016$). What does
not separate the sets is also informative: the first-to-fourth gap
distinguishes replicators only across families, not within them ($p = 0.21$),
the first-to-second gap separates them in neither test,
and popularity does not distinguish them at all ($p = 0.14$): nothing here
marks the replicators as obscure corners of the panel.

Three caveats apply. The effects are modest (roughly 62--65\%
of replicator--base pairs ordered as predicted). The same measures do not predict
skew size across ordinary members: sharpness picks out the high-skew tail but
does not grade the panel, and the quiet row of Figure~\ref{fig:boards}
shows replicated skews with no tactical signature at all. And the
replication threshold is outcome-derived, so
the comparison is descriptive and post hoc. It is consistent with, but does not
establish, the reading that such structures are intrinsically harder for one side
to play in practice (Section~\ref{sec:scope}).

\subsection{What the clock burden does and does not establish}\label{sec:clockscope}
The clock result moves the skew from the scoresheet to the board. Without it,
the skew is an accounting fact about millions of finished games; with it, the
deficit is visible in play: at positions the engine scores as level, one
side reliably stops to think while the other moves on. And it is
substantially the same players thinking longer whenever they sit on a
disfavoured side. Two objections an informed reader should raise bound what
the result can mean.

The first is clock economics: in fast chess a long think is itself costly, so
perhaps the disfavoured side loses \emph{because} it spends clock, or
habitually slow players lose wherever they sit. The evidence resists that
reading. The disjoint-halves re-fit removes the
mechanical version: no shared game links a think time to the skew it
predicts. The within-player retention removes the compositional version: the
asymmetry is not carried by slow players populating the disfavoured sides. A
pure clock-cost mechanism should also strengthen as the clock shortens, and
it does not: the coefficient is flat from blitz to classical, and among
classical occurrences (0.9\% of the data, but a precisely estimated
slice, where a think of this size is nearly free) the same asymmetry
tracks a skew map estimated overwhelmingly from faster games. What this
analysis cannot do is bound the cumulative clock consequence: the
measured cost is about one standardised second against matched controls at
plies 2--20, too little on its own to decide a game, but
think-time differences later in the game are unmeasured, so the reverse
channel is constrained, not excluded. On this evidence the long think is
best read as a marker of the deficit, not as its cause at the position
itself.

The second objection is more serious: the favoured
side is not merely spared the long think. It moves faster than
neutral-position controls, the signature of one side replying
from home preparation while the other is out of book. This is the
repertoire-selection confound in behavioural form, and this design cannot
separate it from intrinsic difficulty: both predict the same one-sided time
cost, and within-player demeaning removes a player's general speed, not that
player's line-specific preparation. Both readings leave the same
finding intact: at positions the engine calls equal, one side
reliably pays a practical, at-the-board cost that the evaluation cannot see.
Whether that cost originates in the position or in how humans prepare for it
is the causal question (Section~\ref{sec:scope}). Either way, the engine's
zero describes the board, not the contest.

The measurement itself has limits. Think time is measured for a single move,
not integrated over the game, and is a noisy proxy for difficulty: a
player can be lost without noticing, and spend nothing. There is no measure
of move \emph{quality}, so whether the disfavoured side also errs more
remains open. And time pressure in the narrow sense is untestable here:
fewer than one occurrence in 50{,}000 has the mover below 10\% of the base
clock, because these positions occur by ply 20.

\subsection{Scope and limitations}\label{sec:scope}
\paragraph{The result is observational and non-causal.}
We have not shown that the position causes the skew. Within an ECO code there
can be narrower sub-repertoires with systematically different player pools, so a
positive within-ECO replication slope can still reflect selection on who plays a line rather
than the difficulty of the position itself. Our within-family fixed effects remove the
family-level version of this confound but not the sub-family version. That residual is
the boundary of the claim. Concretely, the leading mechanisms that could
live inside the residual are line-specific \emph{preparation} (players entering
lines they have studied); \emph{player-by-line specialisation} beyond
preparation (repertoire comfort); \emph{transposition history} (different move
orders selecting different populations into the same node); and
\emph{multi-accounting or engine assistance} surviving platform moderation.
Each is compatible with every replication in this paper, and none is
distinguishable from practical difficulty by this design. The estimand is therefore the
\emph{naturally-reached position}, not a context-free position effect. A causal claim
would require randomly assigning players to play both sides of a position, with
the assigned-game outcome as ground truth --- the design of a pre-registered
companion study, outside the scope of this paper.

\paragraph{Other limitations.}
The demonstrated ecology is online fast chess on one platform: 76\% of analysed
occurrences are blitz, the classical stratum is too thin for a replication
estimate of its own (Section~\ref{sec:robustness}), and the population is Lichess accounts rated
$\ge 1000$. Whether the skews persist over the board or at long time controls
is untested here.
The estimand covers high-occurrence common positions (discovery count $\ge 1000$).
Generalisation to rarer positions is out of scope here. The panel is also
built on the fixed stride-2 sample of the game stream
(Section~\ref{sec:data}): games at the alternate offset were never parsed,
and sensitivity to that choice is untested.

\section{Conclusion}
Among the chess opening positions a strong engine assesses as equal at high
search depth, and that
humans actually reach in sufficient numbers, human results are not balanced.
Positions carry stable, rating-adjusted, account-disjoint outcome
skews (favouring White in some positions, Black in others) that
reproduce across player-disjoint, temporal, and cross-rating-band
re-partitions, within opening family throughout and at both family
granularities, with a realised replication slope of 0.69 on the $\ge 1000$
panel. The existence of the skew is invariant to every
adequately powered perturbation of the position set and the analysis design we tested
(Table~\ref{tab:robust}, Section~\ref{sec:robustness}), while the replication
slope's numeric value is a property of the panel mix.
Nor is the pattern specific to the month analysed: the
per-position skews measured in October 2025 reproduce on June 2026 games with a
cross-month replication slope of 0.90.
The skew also has a counterpart at the board: at these positions, the
disfavoured side reliably spends more clock time
(Section~\ref{sec:clockscope}).
The result is observational: it establishes that the skew is real and not an
artefact of rating gaps, individual players, or family-level opening choice.
Selection into lines within a family remains part of the measured estimand,
and whether the position itself is the cause requires a randomised assigned
play-out, the subject of a pre-registered companion study.
We release the deep-verified position set and per-position skew estimates so the
existence result can be independently verified and extended.

\clearpage
\appendix
\renewcommand{\thetable}{A\arabic{table}}
\setcounter{table}{0}
\renewcommand{\thefigure}{A\arabic{figure}}
\setcounter{figure}{0}
\section{Split construction}\label{sec:splits}
Table~\ref{tab:splits} documents each data re-partition's rule and realised cells.
The primary cells are account groups spanning the whole month. The calendar halves
of Section~\ref{sec:data} serve position selection and the temporal axis. The
primary split retains 49.97\% of usable occurrences (games whose players fall in
different groups are dropped), and the group assignment is exactly reproducible
from the pinned seed in the released artefact.

\begin{table}[!htb]
\centering
\footnotesize
\setlength{\tabcolsep}{4pt}
\caption{Split construction by replication axis: partition rule and realised
cells. Occurrence counts are usable rows. Group assignment is a uniform random
permutation of the casefolded account ids (seed pinned in the released artefact).
A game is kept only when both accounts share a group. Calibration is fitted on
each axis's first-listed cell and applied, unchanged, to the second. Estimator
samples: primary 1{,}630 positions in 118 ECO-code families, 1{,}661 in 5
ECO-letter families, temporal 1{,}628 in 118.}
\label{tab:splits}
\begin{tabular}{l >{\raggedright\arraybackslash}p{5.6cm} r r r}
\toprule
Axis: cell & Partition rule & Occurrences & Games & Accounts \\
\midrule
Primary: group A & permutation group A (572{,}360 ids); both accounts in A &
4{,}039{,}025 & 2{,}248{,}684 & 484{,}109 \\
Primary: group B & permutation group B (572{,}361 ids); both accounts in B &
4{,}024{,}113 & 2{,}243{,}636 & 484{,}404 \\
Temporal: early & October 1--15; all games &
7{,}696{,}273 & 4{,}289{,}795 & shared\textsuperscript{a} \\
Temporal: late & October 16--31; all games &
8{,}438{,}660 & 4{,}695{,}530 & shared\textsuperscript{a} \\
Bands: L / M / H & per-game mean rating $<$1500 / 1500--1800 / $\ge$1800;
floors 100 per band, 50 per cell & \multicolumn{3}{c}{pair samples 1{,}535 /
1{,}604 / 1{,}514\textsuperscript{b}} \\
\bottomrule
\end{tabular}

\medskip
\begin{minipage}{0.97\linewidth}
\raggedright\footnotesize
\textsuperscript{a}\,The temporal cells do not block on accounts. The same
account can appear in both halves.
\textsuperscript{b}\,Band cells are per-game strata. An account can contribute
to more than one band (Section~\ref{sec:results}). Pair samples are the
positions entering the cross-band fit (families with $\ge 2$ members), in the
order L--M / M--H / L--H.
\end{minipage}
\end{table}

\section{Estimator and measurement details}\label{app:methods}

\subsection{Calibration specification}\label{app:calib}
The expected-score model (Section~\ref{sec:calib}) is a fractional-logit GLM
\citep{papke1996fractional}: the realised White score is regressed on a natural
cubic spline in the rating difference (4 degrees of freedom), the centred mean
rating of the two players, their interaction, time-control indicators, and
time-control $\times$ rating-difference interactions, fitted by quasi-maximum
likelihood on the axis's first-cell occurrence rows. The colour-aware refit of
Table~\ref{tab:robust} adds a side-to-move indicator and its rating-difference
interaction to this specification.

\subsection{Effective sample size and reliability}\label{app:neff}
For one position and one colour, let $c_a$ be the number of its retained games
in which account $a$ plays that colour; the Kish count is
$(\sum_a c_a)^2 / \sum_a c_a^2$, and $n_{\text{eff}}$ is the smaller of the
White-side and Black-side counts. Each position's sampling variance is the
one-way cluster-robust variance of its mean residual, clustered by account,
taking the larger of the White- and Black-clustered values. The reliability
(ICC) quoted in Table~\ref{tab:headline} is
$(V_{\text{obs}} - \bar V_{\text{samp}})/V_{\text{obs}}$, clamped to $[0,1]$,
where $V_{\text{obs}}$ is the $n_{\text{eff}}$-weighted variance of the
family-demeaned anchor-cell skews and $\bar V_{\text{samp}}$ the weighted mean
sampling variance. The disattenuated slope is $\hat{\beta}$ divided by this
reliability, and a disattenuated interval divides the raw-slope interval by
the same point ICC, treating it as known: the ICC's own sampling uncertainty
is not propagated, so the disattenuated interval is indicative rather than
fully propagated. An account-split half-sample correlation cross-checks the
reliability estimate.

\subsection{Permutation-test calibration}\label{app:permcal}
The calibration check of Section~\ref{sec:gradient} simulates a null with no
position effect: each position's group-B skew is redrawn as its
precision-weighted family mean plus $\mathcal{N}(0,
\hat{\sigma}^2/n^{B}_{\text{eff},i})$ noise, with the group-A skews and the
full precision profile held at their observed values (the observed
$n_{\text{eff}}$ spread is 167:1). Over 1{,}000 simulated panels $\times$ 199
permutation draws each, the test rejects at 0.047 and 0.013 against nominal
0.05 and 0.01. The studentised variant (each side demeaned against its
precision-weighted family mean, then divided by its sampling standard error)
is likewise calibrated (0.048 / 0.011). An unpaired variant that shuffles
$\delta^{B}$ alone, leaving each receiving position's weight in place,
inflates the null spread (SD 0.036 vs.\ 0.028) and is conservative rather than
liberal (rejection 0.005 at nominal 0.05): the pairing protects power rather
than validity.

\subsection{Within-account outcome demeaning}\label{app:withinplayer}
Constructed from the occurrence dataset of the primary split (8{,}063{,}138
rows; 968{,}513 accounts, one identity per account across both colours). For
each account an oriented mean residual is computed: the mean of its games'
calibration residuals, taken from White's perspective when the account held
White and negated when it held Black. Each occurrence is then demeaned two
ways with leave-one-out means,
$\tilde r = r - \bar a_{\mathrm{loo}}(\text{White account}) + \bar a_{\mathrm{loo}}(\text{Black account})$,
and occurrences in which either account appears only once are dropped
(1.6\% of rows). Skews and the replication slope are recomputed from
$\tilde r$ by the unchanged estimator, with the plain slope re-fitted on the
identical rows and weights for a like-for-like comparison. Inference is the
family-block permutation of Section~\ref{sec:gradient}. Because the split is
player-disjoint, account means never mix its two cells. The leave-one-out
unit is the occurrence: the current row's own contribution is removed
exactly, while occurrences of the same game at other member positions remain
in the account mean, so a residual own-game term survives in proportion to
same-game duplication (additional attenuation of the within slope, in the
conservative direction for the existence claim, and not included in the
factor below). The sampling noise the demeaning leaves in the group-A skews
(the regressor of the fit; noise on the group-B side widens intervals
without attenuating) implies a mechanical attenuation factor of 0.955, so
of the observed within-to-plain ratio of 0.78, at least 0.05 is mechanical,
and the remainder is an upper bound on account-level outcome structure,
since account means also absorb genuine position skew through an account's
repeated play of particular lines. Demeaning removes roughly a third of the
within-family skew variance in both halves.

\subsection{Clock-burden construction}\label{app:clock}
\paragraph{Retention.}
Of the month's 16{,}135{,}028 occurrences, 442{,}530 lack a mover clock because
they sit at ply 1, where no preceding clock comment exists to read the mover's
clock from (structural rather than data loss, costing exactly one
position, the initial one); 3{,}617 are increment-only time controls (no base
clock, so clock fractions are undefined); 1{,}036 carry physically impossible
negative reconstructed think times; and 9 think times exceeding the base clock
are winsorised to it. Retained: 15{,}687{,}845 occurrences (97.2\%) at 1{,}660
of the 1{,}661 positions.

\paragraph{Think time and standardisation.}
The think time for the move played from the position is
$t = \text{clock}_{\text{before}} - \text{clock}_{\text{after}} +
\text{increment}$. Raw seconds are incomparable across clock formats, so
$\log(1+t)$ is standardised within cells defined by (time control, base
seconds, increment seconds, ply):
$z = (\log(1+t) - \mu_{\text{cell}})/\sigma_{\text{cell}}$. Cells with fewer
than 100 occurrences are dropped (99.6\% of occurrences retained). The
position-level outcome is the occurrence-mean of $z$.

\paragraph{Regression.}
The fitted model regresses the position-level mean $z$ on the
mover-point-of-view skew $\delta_{\text{mover}}$ ($+\delta$ when White is to
move, $-\delta$ when Black is; $\delta$ is the precision-weighted mean of the
two account-split estimates), with opening-family fixed effects and controls
for $|\delta|$, side to move, ply, and mover rating, weighted by occurrence
count, with family-clustered standard errors (1{,}629 positions in 118
families; 15.5M occurrences). The side-to-move control is required because
$\delta_{\text{mover}}$ flips sign with colour. Its own coefficient is
$\approx 0$.
Section~\ref{sec:results} quotes two further re-fits. In the cross-split
re-fit, each position's mean think time is computed from one calendar half of
the games and its skew $\delta$ from the other half, so no game contributes
to both sides of the regression. The two directions give $-3.38$ and $-3.53$.
In the within-player re-fit, each think time is first demeaned by that
player's own average think time over all member positions (computed leaving
the current occurrence out), so anything that is merely ``this player is
slow'' drops out. The coefficient is $-2.99$.

\paragraph{Interpretable units.}
The occurrence-weighted mean cell standard deviation of $\log(1+t)$ is 0.593,
so $\hat{\beta}_{\text{signed}} = -3.54$ implies that a 0.10 swing in
mover-POV skew (favoured $+0.05$ to disfavoured $-0.05$) shifts think time by
0.354 cell standard deviations: a factor of $\approx 1.23$ in $1+t$, the
``23\% more think time'' (a ratio in $1+t$; back-transformed at the observed
standardised baselines of 2.5--4.2 seconds, the ratio in $t$ itself is
1.28--1.32, so the quoted figure is conservative). The ``67\% longer'' compares the replicated
positions' disfavoured-mover and favoured-mover groups, each reweighted to the
pooled replicator cell distribution so the time-control, increment, and
move-number mix is held fixed: standardised mean think times 4.20 vs.\ 2.51
seconds. A ``long think'' is defined per cell as exceeding the threshold, from
the grid $\{5, 10, 15, 20, 30, 45, 60\}$ seconds, whose pooled exceedance rate
in that cell is closest to 10\%. Against family-matched low-$|\delta|$
controls ($|\bar\delta| < 0.02$, same ECO-code family, at least two controls
required), the disfavoured movers' long-think rate is 12.0\% against the
favoured movers' 5.3\%, the factor 2.3.

\paragraph{Inference and status.}
The permutation $p$ uses 2{,}000 within-family draws under two variants
(shuffling $\delta$ among each family's positions, and flipping the sign of
$\delta_{\text{mover}}$), both at the floor $p = 0.0005$. The
$\delta$-shuffle null centres below zero ($-1.06$) because a within-family
shuffle preserves the sign $\times$ family-mean-skew component. Adding that
component as a nuisance covariate centres the null ($+0.08$) and leaves the
conclusion unchanged ($\hat{\beta}_{\text{signed}} = -2.95$, $p = 0.0005$), so
the reported null is conservative. The clock analysis is a secondary analysis,
specified after the primary outcome results were known. Its
retention rules were fixed from a diagnostic scan before any regression was
run, and the headline coefficient is reported without errors-in-variables
correction (estimated reliability of $\bar\delta$ 0.84; corrected coefficient
$-4.22$).

\subsection{Covariate-matched placebo construction}\label{app:placebo}
The placebo of Section~\ref{sec:robustness} asks what the estimator would
recover from covariate composition alone. Occurrence rows are binned into
cells of time control $\times$ mean-rating quintile $\times$
rating-difference quintile (global quantile bins, fixed once). Within each
(opening family, account group, cell) pool, the rows are randomly re-dealt to
\emph{pseudo-positions} in exactly the real positions' per-cell counts. Each
pseudo-position therefore reproduces its real counterpart's occurrence count
and joint covariate distribution exactly, while its rows are drawn from the
whole family's cell-matched pool, so position identity is destroyed. The two
account groups are dealt independently, so a pseudo-position's two skews
share nothing but the covariate profile, which is the channel a
calibration-misspecification artefact would use. The unchanged estimator
(weights, demeaning, and family fixed effects recomputed from the dealt
rows) is then run per draw. The quoted 0.046 is the mean over seven draws
(range 0.005--0.084), and a reshuffle with no covariate matching at all
returns 0.006. The positive control adds a ply quartile to the cell
definition: ply is a near-identifier of position (the expected fraction of
rows dealt back to their own position rises from 0.21 to 0.38), and the
placebo partially rebuilds the slope (mean 0.350), so the design has power
to detect identity-driven co-skew. The companion covariate-means re-fit adds
to the headline regression, for each account group separately, the
position's mean centred rating level, mean rating difference, mean absolute
rating difference, mean ply, and blitz and classical occurrence shares
(twelve regressors in all).

\subsection{Line-cluster construction}\label{app:nesting}
For each pair of member positions, the \emph{conditional overlap} is the
number of games reaching both divided by the number reaching the rarer of
the two, computed in one pass over the occurrence rows. Only 10{,}594 of the
1{,}378{,}630 possible pairs (0.77\%) co-occur in any game, and the typical
co-occurring pair is nearly disjoint (median Jaccard similarity 0.004), but
the tail is totally nested: 412 pairs share at least 99\% of the rarer
position's games, 177 share all of them, and 1{,}108 of the 1{,}661
positions (66.7\%) lie on an ancestor--descendant chain (conditional overlap
$\ge 0.50$ with a consistent ply order in at least 95\% of shared games).
Line-clusters are the connected components of the graph joining pairs at
conditional overlap $\ge 0.50$ (814 clusters, against 501 at threshold
0.25 and 1{,}072 at 0.75), and the retained representative is the
cluster's highest-$n_{\text{eff}}$ position. Choosing a random
representative instead (five draws per threshold) lowers the slope to
0.57--0.66 (ordinary attenuation from retaining lower-weight, noisier
positions, not a nesting effect), with every re-fit at the permutation
floor. The 71-position replicated set occupies 46 distinct clusters at the
0.50 threshold (36 at 0.25, 53 at 0.75). The four strongest replicators
merge to three at every threshold, the two Traxler positions always joining.

\subsection{Sharpness measures}\label{app:sharpness}
The six measures of Section~\ref{sec:highskew} are computed per position
from the stored depth-ladder MultiPV-4 analysis, with the ``sharper''
direction fixed before testing: (i) the depth-28 \emph{spread}, the
evaluation gap between the engine's first and fourth candidate moves
(larger = sharper); (ii) the \emph{only-move margin}, the gap between the
first and second candidates (larger); (iii) the \emph{principal-variation
entropy}, the entropy in bits of a softmax (temperature 100\,cp) over the
four depth-28 candidate evaluations (lower = the advantage concentrated in
one move); (iv) the standard deviation and (v) the range of the
White-point-of-view evaluation across the depth ladder $(12,16,20,25,28)$
(larger = less settled); and (vi) the \emph{terminal drift}, the absolute
evaluation change over the last rung, depth 25 to 28 (larger). The first
three measure how concentrated the position's resources are in a single
move. The last three measure how settled the search is. The test statistic is the
mean family-demeaned replicator-minus-base difference. The replicator label
is shuffled within ECO-code family (1{,}000 draws), so the comparison runs
inside the 31 families containing both classes (1{,}013 positions, 62 of
the 71 replicators). Cliff's $\delta$ is reported as the marginal effect
size, and the Benjamini--Hochberg correction runs across the six measures.
Beyond the three measures that separate the sets
(Section~\ref{sec:highskew}), the only-move margin and the terminal drift
do not ($p \ge 0.23$), and the depth-28 spread separates only across
families, not within them.

\section*{Data and code availability}
The datasets and analysis code supporting the conclusions of this article are
archived in a public Zenodo archive with a persistent DOI:
\href{https://doi.org/10.5281/zenodo.21629354}{10.5281/zenodo.21629354}.
The archive holds the analysis
code with its hermetic test suite; the deep-verified \sfzero{} position set
(1{,}661 member positions, as full FEN strings and position hashes); the
per-position skew table (FEN, the group-A/group-B and calendar-half $\delta$
pairs, $n_{\text{eff}}$, ECO family); the complete robustness-check result
files behind Table~\ref{tab:robust}; and the run manifest recording every
pinned parameter, stage command, realised count, and SHA-256 content
fingerprint needed for independent verification and end-to-end reproduction.
The archived derived data are position-level and contain no account
identifiers, direct or coded. Pseudonymous account names are used only
during analysis. The Zenodo deposit is the identifier of record. A working copy of the code is
additionally available on GitHub
(\url{https://github.com/jesung/engine-equal-human-unequal}). The
third-party Lichess database dumps we
analyse (October 2025, and June 2026 for the external-month check) are released
under CC0 and are identified by their pinned digests in the run manifest, not
redistributed.

\section*{Acknowledgements}
The author thanks the Lichess team for maintaining and openly releasing the game
database, and the Stockfish and Leela Chess Zero communities for their open-source
engines.

\bibliographystyle{plainnat}
\bibliography{refs}

\end{document}